# Cancer Subtype Identification through Integrating Inter and Intra Dataset Relationships in Multi-Omics Data

M. Peelen[1], L. Bagheriye[2], Member, IEEE and J. Kwisthout[3]

[1,2,3]Donders Institute for Brain, Cognition and Behaviour, Radboud University Nijmegen, P.O. Box 9104, 6500HE Nijmegen, Netherlands.

Corresponding author: L.Bagheriye Author (Leila.Bagheriye@donders.ru.nl).

**ABSTRACT** The integration of multi-omics data has emerged as a promising approach for gaining comprehensive insights into complex diseases such as cancer. This paper proposes a novel approach to identify cancer subtypes through the integration of multi-omics data for clustering. The proposed method, named LIDAF utilises affinity matrices based on linear relationships between and within different omics datasets (Linear Inter and Intra Dataset Affinity Fusion (LIDAF)). Canonical Correlation Analysis is in this paper employed to create distance matrices based on Euclidean distances between canonical variates. The distance matrices are converted to affinity matrices and those are fused in a three-step process. The proposed LIDAF addresses the limitations of the existing method resulting in improvement of clustering performance as measured by the Adjusted Rand Index and the Normalized Mutual Information score. Moreover, our proposed LIDAF approach demonstrates a notable enhancement in 50% of the $\log_{10}$ rank p-values obtained from Cox survival analysis, surpassing the performance of the best reported method, highlighting its potential of identifying distinct cancer subtypes.

## I. INTRODUCTION

Cancer subtype diagnosis is a field of critical importance, as accurate identification empowers medical professionals to make informed decisions regarding the most suitable treatment options, thereby maximizing the likelihood of achieving favourable patient outcomes [1]. Over the past two decades, significant efforts have been made to identify cancer subtypes through, for example, the exploration of gene expressions and genomics in cancer cells, which have provided valuable insights into studying organisms at the DNA and RNA levels [2]. To gather this information and mitigate the challenges arising from the disparity between the number of samples and the high dimensionality of features, multiple omics datasets of the same type are merged to generate a single omics dataset. This, however, introduces heterogeneity and more complexity in the data, which could invoke the "Curse of Dimensionality" [3].

For this paper, omics data is obtained from The Cancer Genome Atlas (TCGA), which is a comprehensive source that encompasses gene expressions, miRNA, and DNA methylation, spanning across various cancer types [4-7].

Gene expressions includes protein-coding genes, non-coding RNA genes and pseudogenes [8]. Analysing this type of data provides insight into the dysregulation of genes associated with specific types of cancer. The strength of the dataset lies in the inclusion of differentially expressed genes, which improves the reliability of gene identification and allows a better understanding of molecular changes associated with specific types of cancer.

Another omics type used in this paper is miRNA. miRNA is a small type of RNA that does not encode proteins but has a significant impact on gene expressions. In the context of cancer, miRNA expressions are often impaired. This disruption can occur through several mechanisms, including amplification or deletion of miRNA genes, irregular control of transcription, or disruptions in epigenetic modifications [9]. As a result, miRNA shows potential as a biomarker for clinical applications.

DNA methylation represents the third type of omics utilized in this paper. It serves as an important mechanism playing a crucial role in regulating gene expressions in both normal and cancer cells [10]. Dysregulation of DNA methylation can lead to inappropriate silencing of tumour suppressor genes or the activation of oncogenes, contributing to cancer development and progression.

Initially, most approaches for identifying cancer subtypes relied on distinct biological characteristics or clinical outcomes, but were constrained by using only a single omics type. As a result, these methods struggled to provide a comprehensive description of the organism's clinical information and fell short in capturing the subtle complexities of cancer [11-18]. To address this issue and unveil the molecular mechanisms governing specific biological processes and interactions, using multiple omics type datasets simultaneously will result in a more comprehensive approach and is referred to as "multi-omics data" [19]. However, the utilization of multi-omics data increases the complexity and heterogeneity of the data when compared to using a single omics dataset [3,20-22].

To address the challenges arising from this complexity and heterogeneity, several studies have explored various approaches aimed at overcoming these obstacles. Among this research, the current leading method is the multi-omics data integration for clustering to identify cancer subtypes (MDICC) framework [23]. It fuses affinity matrices by selecting and weighting heterogeneous data sources, allowing for clustering and subsequent survival analysis to aid in the identification of cancer subtypes. This method has certain limitations, including the utilization of all features in the dataset and the lack of incorporation of the relationship between different omics types.

In response to these limitations, this paper proposes the following research question:

*Does the addition of affinity matrices to the data integration of multi-omics data, based on the relationship between omics datasets, improve clustering performance and $-log_{10}$ rank p-values compared to four state-of-the-art clustering methods?*

To investigate this research question, the following null hypothesis ($H_0$) and alternative hypothesis ($H_A$) can be formulated:

- **$H_0$:** The addition of affinity matrices to the data integration of multi-omics data, based on the relationships between different omics types using pre-processed data, does not result in improved clustering performance and does not lead to higher significant survival $-log_{10}$ rank p-values compared to four state-of-the-art clustering methods.
- **$H_A$:** The addition of affinity matrices to the data integration of multi-omics data, based on the relationships between different omics types using pre-processed data, improves clustering performance and leads to higher significant survival $-log_{10}$ rank p-values compared to four state-of-the-art clustering methods.

To reject the null hypothesis and accept the alternative hypothesis, this paper proposes a comprehensive approach. The proposed approach encompasses several data pre-processing techniques, including the removal of missing values and zeros, value imputation, Z-score standardization, variable selection using Gaussian Mixture Model with Bayesian Inference, and the identification of linear relationships through Canonical Correlation Analysis (CCA). The objective of this method is to select relevant features and utilize CCA to uncover relationships between different omics types introduced earlier in this paper. These relationships are used to create affinity matrices, which are then incorporated with the affinity matrices of [23] through a three-time fusion process.

Our proposed method which is utilizing Linear Inter and Intra Dataset Affinity Fusion (LIDAF) provides evidence to reject the null hypothesis suggesting that it has the potential to improve clustering performance across all four datasets based on the Adjusted Rand Index (ARI) and Normalized Mutual Information (NMI) score and our proposed method has the highest number of p-values with greater significance compared to four other state-of-the-art clustering methods. From this it can be concluded that LIDAF can address the limitations [23] with almost the same or even better performance.

The subsequent sections of this paper are structured as follows. Section II provides a comprehensive review of related works that diagnose cancer subtypes based on genomics from TCGA as presented in Table I. In section III, the proposed LIDAF is elaborated upon, outlining its key aspects and methodology. Section IV presents the results and analysis. Finally, section V concludes the paper, summarizing the key findings and implications of this study and section VI addresses the limitations and possibilities for future work.

## II. THEORETICAL BACKGROUND

The introduced proposed method aims to enhance the clustering performance and statistical significance in survival analysis, thereby providing significant prove that the method helps with identifying cancer subtypes. By incorporating LIDAF into the existing literature, efforts are made to tackle the limitations of integrating multi-omics data [23]. The following sections will present techniques aimed at addressing the challenges posed by sparse, complex, and heterogeneous omics data. The objective is to provide a comprehensive understanding of the challenges involved and explore the potential enhancements offered by the proposed LIDAF.

### A. MISSING VALUES

Integrating multiple studies on a particular omics type into a single dataset has the potential to enhance the sample-to-feature ratio. However, this integration often leads to a lack of consideration for the same features across different studies, as each study may prioritize different features within the omics type [3,20,21]. Consequently, the datasets presented in Table I may contain missing values, often denoted as zero. Hence, the consideration of the nature of zeros is beneficial to mitigate the sparsity that could introduce bias, noise, computational challenges, and the reduction of information [24-26]. This nature can be differentiated between biological zeros and non-biological zeros [27].

Biological zeros are zeros that accurately represent the absence of a particular variable in the biological system under study. These zeros reflect genuine biological characteristics and should not be removed or altered during data analysis. In contrast, non-biological zeros can further be classified into two subcategories: technical zeros and sample zeros [26]. Technical zeros arise during the library preparation process, which involves enzymatic and chemical reactions on DNA or RNA. In some cases, the values resulting from these reactions may fall below a predefined threshold, resulting in the assignment of a value of zero. These technical zeros are considered non-biological and are typically a result of experimental procedure rather than a true biological absence. On the other hand, sample zeros, occur due to incomplete or incorrect measurements of omics data for certain samples. These zeros can result from various factors, such as technical limitations or specific characteristics of the sample themselves. To address the issue of non-biological zeros it is important to establish a threshold for their exclusion from the analysis. Features containing over 20% zeros may be removed from the analysis, considering the computational costs of further analysis and the possibility of them being non-biological zeros [26, 28-30].

After removal of features containing more than 20% zeros, it is possible that sparsity remains in the data for observations with fewer zeros. To address this issue, imputation methods can be utilized to estimate values for the missing entries. One such imputation method is the classical machine learning approach known as K-Nearest Neighbour (KNN) imputer, which offers the advantage of a relatively straightforward implementation [31]. However, it may suffer from lower accuracy or present challenges in obtaining a valid value for $K$.

*B. DATA PREPARATION*
There are several strategies available for preparing omics datasets for comparison. One notable approach, as highlighted by [32], involves the use of Z-score standardization. This technique centres, scales, and standardizes omics data to enhance performance and complexity. This approach proves valuable in addressing the challenges associated with omics data when compared to other preparation methods [33,34]. By normalizing the data, the Z-score standardization enables meaningful comparisons across different omics datasets, eliminating scale differences and ensuring equal weight for different features. This normalization step facilitates the detection of patterns, accounts for zero mean and equal variances, and aligns with assumptions made by many statistical and machine learning methods such as CCA [35-37]. While there are other methods capable of achieving good results, the advantage of the Z-score standardization lies in its ease of adaptability while maintaining a rigorous approach and minimizing arbitrary outcomes when compared to other methods [33]. Normalization through Z-score standardization increases the visibility of outliers and provides equal weights to the data. In cases where those outliers remain prominent, applying a transformation can be beneficial. Besides handling outliers, such a transformation can also help with achieving a more symmetric distribution of skewed data. This is particularly useful for addressing heteroscedasticity, as excessive heteroscedasticity can introduce bias, violate statistical assumptions, and undermine the reliability of results [38]. A commonly used transformation method in omics data analysis is the logarithmic transformation [39]. It is effective in handling data that follows an exponential distribution and can improve interpretability of data. However, there is a challenge when dealing with negative values. Negative values cannot be directly transformed using logarithms. In the context of Z-score standardization, negative values could arise during the scaling step when a data point is smaller than the variable's mean. To address this issue, a pseudocount is sometimes added to negative values. However, recent studies have highlighted that inclusion of such a pseudocount may introduce distortion in data [39-41]. Therefore, another technique can be used such as a power transformation. An example is the Box-Cox transformation and is represented by [42]:

$$x_{i,j}^{(\lambda)} = \begin{cases} \frac{x_{i,j}^{\lambda}-1}{\lambda} & \text{if } \lambda \neq 0, \\ \log(x_{i,j}) & \text{if } \lambda = 0, \end{cases} \quad (1)$$

where $\lambda$ is the power of a datapoint $x_{i,j}$ and is determined by a value that maximizes the likelihood of achieving normality and equal variance in the data. Equation 1 could improve distributional properties in data and address any heteroscedasticity concerns. However, it also has the limitation that it cannot handle negative values. To address this limitation, an alternative power transformation called the Yeo-Johnson transformation, based on the Box-Cox transformation in Equation 1, can be used and is represented by [44]:

$$x_{i,j}^{(\lambda)} = \begin{cases} \frac{(x_{i,j}+1)^{\lambda}-1}{\lambda}, & \text{if } \lambda \neq 0, x_{i,j} \geq 0, \\ \log(x_{i,j}+1), & \text{if } \lambda = 0, x_{i,j} \geq 0, \\ \frac{-(-x_{i,j}+1)^{(2-\lambda)}-1}{2-\lambda}, & \text{if } \lambda \neq 2, x_{i,j} < 0, \\ -\log(-x_{i,j}+1), & \text{if } \lambda = 2, x_{i,j} < 0, \end{cases} \quad (2)$$

The main difference between Equation 2 and Equation 1 lies in the last two cases of Equation 2. These cases differentiate between the value $\lambda \neq 2$ and $\lambda = 2$. In the latter case, the Yeo-Johnson transformation focuses solely on handling negative values since it utilizes the natural logarithm. This is accomplished by negating the negative variable and adding 1 to appropriately handle zero values. For $\lambda \neq 2$, the Yeo-Johnson transformation introduces a negative sign to preserve the correct direction of the data. To make these properties effective, it is necessary to assign a specific $\lambda$ for each variable. Determining the optimal $\lambda$ by using the maximum likelihood can be debatable [44,45]. For instance, the maximum likelihood can be sensitive to outliers, whereas the transformation should reduce the impact of outliers. However, other transformation methods are also commonly susceptible to extreme values and rely on them to a considerable degree [46].

*C. RELATED METHODS*
The preceding sections have discussed various aspects related to omics data to identify cancer subtypes. Nevertheless, there exist alternative methodologies for discerning cancer subtypes. For instance, one such approach involves the utilization of Fiber-optic Surface Plasmon Resonance (SPR) sensors at specific wavelengths within cancer cells, outlined in [47]. This method boasts advantages such as reduced costs, simplicity, and the absence of labelling and technical requirements.

Nonetheless, it is worth noting that omics-based approaches offer high accuracy, which is underscoring the importance of exploring strategies to mitigate their inherent challenges. An example of such an approach is Multi-view Subspace Clustering Analysis (MSCA) [72]. MSCA excels at capturing intricate complexities and heterogeneities across multiple views by employing a two-step nonlinear pattern identification for omics data during pattern fusion. Another noteworthy state-of-the-art clustering method for identifying cancer subtypes using omics data is Similarity Network Fusion (SNF) [28]. SNF constructs sample networks and computes their similarities, harnessing the data's complementarity by fusing them together. However, according to [74], the computationally could be further reduced by applying computational optimizations in SNF, leading to the introduction of Affinity Network Fusion (ANF). ANF achieves comparable or superior performance to SNF, while being less computationally heavy, and it also incorporates weights on each view while maintaining a more interpretable outcome.

When it comes to identifying cancer subtypes using omics data MSCA, SNF, and ANF all have a commonality: they rely on multi-omics data. A distinction can be made between the analysis of single-omics data, primarily focused on identifying relevant genes within a specific genome, and the analysis of multi-omics data that incorporates multiple omics datasets. The analysis of multi-omics data involves integrating and considering multiple types of omics data to obtain a comprehensive understanding of biological systems. This approach goes beyond the examination of individual omics datasets and allows for a more holistic exploration of molecular interactions and biological processes [48] due to the dysregulation and complexity of cancer genomes, which are influenced by multiple molecular mechanisms [23]. Utilizing multi-omics data for accurate prediction of cancer subtypes is an area of significant interest. Leveraging multi-omics data holds great potential for improving the understanding of cancer subtypes and advancing precision medicine approaches. The currently leading method accomplishing this, proposed by [23], is MDICC. Unlike other methods that utilizes a spectral clustering approach with a single Gaussian kernel, MDICC constructs affinity matrices with local scales for calculating the Gaussian kernel within the kernel function. This allows considering the distribution of data points within clusters. The inspiration for this approach comes from [49], and [23] utilizes the KNN method to construct a local scale for each node, represented by $j$, in the dataset. The computation of the local scale for each dataset can be performed using the following equation:

$$A_l(x_j, x_n) = \exp\left(\frac{-d^2(x_j, x_n)}{0.5\sigma_j\sigma_n + 0.5d(x_j, x_n)}\right), \quad (3)$$

where $\sigma_j = \frac{\Sigma_{j \in \text{KNN}(x_j)} d(x_j, x_n)}{K}$, and $d(x_j, x_n)$ is the Euclidean distance between sample $j$ and sample $n$ and $K$ represent the top $K$ neighbours of node $j$.

To overcome the noise challenge in omics data, [23] employs a sparse subspace learning framework for the fusion of multiply affinity matrices. This is achieved by minimizing the discrepancy between the fusion network $S$ and the individual affinity matrix $A$, where the importance of each affinity matrix is weighted by the factor $\alpha_l$. To enhance generalization and preserve the low-rank structure of $S$, a Laplacian normalization term is introduced. This term promotes a block-diagonal structure in $S$, where samples within the same subpopulation exhibit high similarities. Constraints and hyperparameters are further employed to enforce sparsity and low-rank properties of the fusion network. To achieve a balanced selection of affinity matrices for fusion, an entropy constraint is introduced on the weighting variable $\alpha_l$. This will result in computing the objective function with updating the parameters as follows [23]:

$$S_{\text{updated}} = \min - \Sigma_l \Sigma_{j,n} \alpha_l A^{(l)}_{j,n} S_{j,n} + 0.5\Sigma_l \alpha_l ||A^{(l)}||_F^2 + \beta||S||_F^2 + \lambda\text{tr}(F^T(I_m - S)F) + \Sigma_l \gamma \alpha_l \log \alpha_l$$

$$(4)$$

$$\text{such that} \begin{cases} F^T F = I_C \\ \Sigma_l \alpha_l = 1, \alpha_l \geq 0 \\ \Sigma_n S_{j,n} = 1, S_{j,n} \geq 0 \end{cases},$$

where $I_n - S$ is the graph Laplacian matrix and $I_m$ the identity matrix, tr is the matrix trace, $\lambda$ is a non-negative parameter and the regularization strength is controlled by the tuning parameters $\beta, \gamma > 0$. These parameters are identical and defined as follows [23]:

$$\gamma = \beta = \frac{1}{N}\Sigma_{j=1}^{N}\Sigma_{n=1}^{k}(s_{j,k+1}^2 - S_{j,n}^2). \quad (5)$$

Finding the optimal solution for Equation 5 can be challenging, because the objective function for variable $S$, $F$, and $\alpha$ exhibit convexity when the remaining variables are held constant. [23] uses a multiplier method to approximate the optimal solution to tackle this problem. The approach involved an iterative optimization procedure where one variable at the time is updated while keeping the remaining variables fixed until convergence. The iterative optimization process led to a progressive enhancement of the objective function.

To examine the performance of the method proposed in [23], Cox proportional hazard regression is used for survival analyses, which uses Equation 6 as hazard function [50]:

$$\lambda(t|X_i) = \lambda_0(t)\exp(\beta \cdot X_i), \quad (6)$$

where *i* is the number of samples and *t* is the time point or duration at which hazard is being measured or evaluated. Moreover, this hazard function and the covariates are used to understand the impact of variables on survival analysis [50]. Clustering labels obtained from the data serve as the covariate, allowing assessment of whether clustering aids in identifying cancer subtypes. The significance of the identifying subtypes is evaluated using the -$\log_{10}$ rank test to obtain the p-value. In all ten cancer datasets used in [23], where p-values for $P \geq 1.30$ is considered as significant, are significant. For clarity, $\log_{10}(0.05)$ is equal to 1.30.

One area of concern is the utilization of all features within the omics data, which may lead to the inclusion of irrelevant features and thereby introducing additional complexity to the analysis [23]. Nevertheless, the method effectively addresses this issue through the integration of a least-weighted square, sparse subspace representation, and entropy methods.

However, the performance of the MDICC method is influenced by certain factors. One limitation is that it primarily focuses on identifying relationships between samples by utilizing affinity matrices that incorporate all attributes. Consequently, it does not fully consider the interaction between different datasets. Additionally, the method solely relies on distances within a single dataset when constructing affinity matrices, overlooking the potential interactions between multiple datasets representing different types of omics.

## III. METHODS

Our proposed LIDAF approach is illustrated in Fig. 1 and described in detail in Algorithm 1. Fig. 1 and Algorithm 1 outline two pipelines: one for finding intra dataset relationships and another for discovering linear inter dataset relationships. Both pipelines take the same three datasets as input. The intra dataset pipeline follows the same preprocessing steps as [23], involving Z-score standardization due to its excellent performance. In contrast, the inter dataset pipeline requires more preparation, involving the removal of missing values based on the 20% criteria of [26, 28-30], followed by KNN imputation. This pipeline proceeds with a Yeo-Johnson transformation to achieve a more normally skewed data distribution for CCA assumptions. To reduce computational costs, a Gaussian Mixture Model (GMM) with Bayesian Inference is used for feature selection. After preprocessing move on to their respective analysis. In the inter dataset pipeline, CCA identifies relationships between different omics types, resulting in six affinity matrices that are fused into one. In the intra dataset pipeline, three affinity matrices are generated and fused into one. This process yields two remaining affinity matrices which are then fused to form the affinity matrix *S*. K-means++ is applied to *S*, followed by survival analysis.

This approach is applied to ten different cancer types, which include: Breast invasive carcinoma (BRCA), Kidney renal clear cell carcinoma (KIRC), Lund adenocarcinoma (LUAD), Liver hepatocellular carcinoma (LIHC), Glioblastoma multiforme (GBM), Lung squamous cell carcinoma (LUSC), Sarcoma (SARC), Ovarian serous cystadenocarcinoma (OV), Acute myeloid leukemia (AML), and Skin cutaneous melanoma (SKCM) [51]. Among these, four cancer types have known true classes, making them applicable for evaluating clustering performance. Furthermore, these ten cancer types have been widely utilized in other studies, facilitating convenient comparison of the results obtained in this paper with those of other methods. LIDAF's primary comparative analysis focuses on the proposed method in contrast to the current leading approach, as referenced in [23], which is adopted as a framework for addressing relationships within omics data. LIDAF introduces a novel pipeline designed to mitigate the constraints observed in MDICC. LIDAF distinguishes itself though data preparation methodologies and a dedicated consideration of the intra relationships among the omics types.

### A. K-NEAREST NEIGHBOUR IMPUTER

First, the data input process is done, using the datasets with the sizes depicted in the third column of Table I. However, to address complexity and mitigate potential bias and noise any feature that has more than 20% missing values or zeros will be removed from the analysis. The removal of missing values in this way aims to enhance the quality and integrity of further analysis without relying on computational methods [26, 28-30]. The resulting dataset sizes can be found in the fourth column of Table I. Subsequently, the remaining missing and zero values were imputed using KNN [28, 52]. The KNN Imputer is a method that imputes missing values by considering the mean value derived from *N* nearest neighbors in the dataset. To determine the proximity between samples, the imputer utilizes the Euclidean distance [53]:

$$d = \sqrt{(x_2 - x_1)^2 + (y_2 - y_1)^2}, \quad (7)$$

where *x* and *y* represent distinct feature values within and between samples, encompassing multiple features. The *K* value, denoted by $K \in [2,N]$, specifies the number of nearest neighbours considered for imputation. In this way, KNN imputer aims to identify similar samples and estimate missing values based on their observed patterns. This approach is instrumental in maintaining the integrity of data relationships, which is a critical requirement within the inter dataset pipeline of LIDAF. However, determining the appropriate N can be challenging. Often this parameter is therefore determined by taking the square root of *N* [54]. For example, in the BRCA dataset with *N* = 323, the value of *K* in KNN imputer will result into 18. Which can be validated by previous studies, such as [29] which also suggested that *K* values for DNA microarrays typically fall within the range of 10 to 20. Fig. 2A shows the right-skewed distribution of the BRCA dataset resulting from the KNN Imputer.

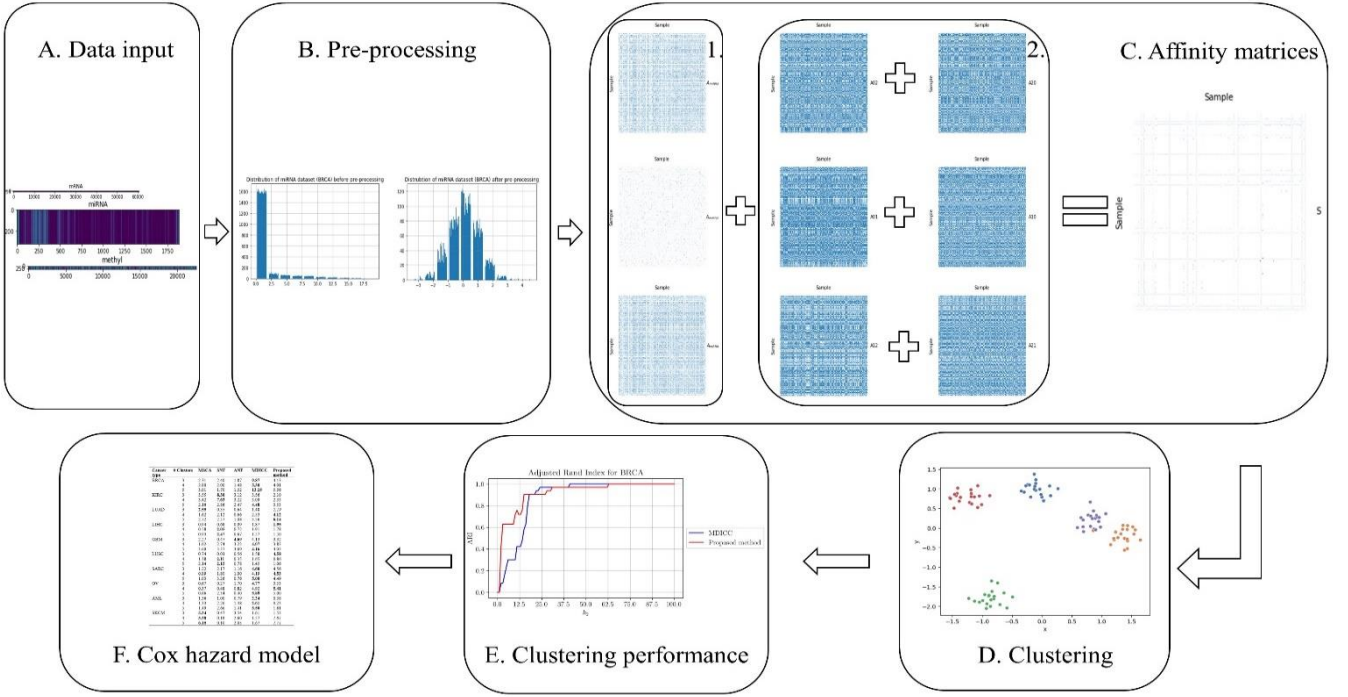

**FIGURE 1.** Overview of the proposed (LIDAF) method. (A) First the data is retrieved; (B) Pre-processing is done, which contains removing missing values, imputing values, standardizing, creating a more normal skewed dataset, and feature selection; (C) that the affinity matrices are generated in C.1 and the affinity matrices in C.2 are created by applying CCA on the pre-processed data. Lastly, C.1 is fused, and C.2 is fused, and those fused matrices are again fused to one affinity matrix S; (D) clustering is done with K-means++; (E)The clustering performance is evaluated; And (F) survival analysis is performed.

### B. NORMALIZATION

The normalization of the proposed LIDAF is done through the Z-score standardization, which can be computed as follows [35]:

$$x_{i,j} = \frac{x_{i,j} - \overline{x_j}}{\sigma}, \qquad (8)$$

where $\sigma$ is the standard deviation as scaling factor and $x_{i,j}$ is a sample. There are two ways to determine $\sigma$. One approach is based on size measures, which are used when the range of values and means of the data differ significantly [35]. Another approach is based on the data dispersion measures, such as unit variance scaling. This approach uses the standard deviation and considers the correlation in the data. In this way, it is ensured that smaller variances are not overshadowed by larger variances, providing a fairer representation of the variable relationship. However, unit variance scaling is sensitive to outliers, which can potentially introduce bias, reduce information, and add noise to the data [32]. To address these potential issues and promote normality in the data, a transformation is applied. Since the standardization pipeline used in our proposed LIDAF method results in negative values, the logarithmic transformation is not suitable. Instead, the Yeo-Johnson transformation is employed. The impact of this transformation is illustrated in Fig. 2B.

### C. VARIABLE SELECTION

The final step in the pre-processing pipeline is variable selection. This step is considered due to a limitation of the MDICC method, namely the use of all features which introduces noise, bias, and complexity [23]. However, due to the subtle interaction present among features in omics data, it is essential to approach feature selection with caution [20]. To do this the GMM with a Bayesian Inference is proposed in the method of this paper since GMM is a commonly used model utilized for variable selection in biometric systems [55]. GMMs modulate data distribution by assuming that observed data originates from a mixture of Gaussian distributions [55, 56]. That is why a transformation is used in the previous pre-processing step. Each component within the mixture corresponds to the distinct cluster or subpopulation present within the data. This approach allows for the identification and selection of relevant features, enabling a comprehensive analysis of the underlying data structure. It could be seen as a weighted sum of M components with Gaussian densities defined as [55]:

$$p(x|\lambda) = \sum_{k=1}^{M} w_k g(x|\mu_k, \Sigma_k), \qquad (9)$$

where $x$ represents a feature as vector, $w_k$ are the weights of $M$ mixtures and the Gaussian densities for $M$ components must satisfy the condition $\sum_{k=1}^{M} w_k = 1$ and looks like [55]:

$$g(x|\mu_k, \Sigma_k) = (1/((2\pi)^{\frac{D}{2}} |\Sigma_k|^{\frac{1}{2}})) \exp(-0.5(x - \mu_k)' \Sigma_k^{-1} (x - \mu_k)), \qquad (10)$$

where $\mu_k$ is the mean vector and $\Sigma_k$ is the covariance matrix. The mean vector represents the average value of the data points belonging to component $k$, whereas the covariance matrix describes the relationship between different features of the data and determines the spread and extent of a Gaussian distribution in each component $k$. There are several ways to assign the spread and Gaussian nature of a distribution. The choice can be made between using a "full", "tied", "diagonal", or "spherical" covariance matrix [52]:

- "*Full*": Each component has its own general covariance matrix;
- "*Tied*": All components share the same general covariance matrix;
- "*Diagonal*": Each component has its own diagonal covariance matrix;
- "*Spherical*": Each component has its own single variance;

The goal of GMM is to estimate $\lambda$, where $\lambda = (w_k, \mu_k, \Sigma_k)$ and $K = 1,\ldots,M$ features. In other words, GMM is parameterized by the mean vectors, covariance matrices and mixture weights. The process to achieve this is by finding the optimal values for the mixture weights, mean vectors, and covariance matrices that best represent the underlying data distribution. Considering the computational costs of pre-processing pipeline, the diagonal covariance matrix is adopted in the variable selection process in the proposed LIDAF [57].

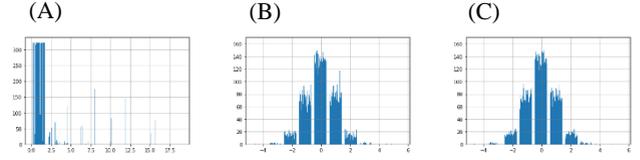

FIGURE 2. Illustration of alterations in the data distribution. (A) Depiction of the original distribution; (B) the distribution after Yeo-Johnson transformation, and (C) distribution after performing feature selection.

The application of GMM can assist in the proper selection of variables for omics data. However, it is accompanied by certain limitations. For instance, this approach frequently requires computational resources, especially when assigning individual covariance matrices to each component. Additionally determining the optimal number of mixture components, to include, remains a subjective task [58,59], as does the selection of the suitable number of features [60].

To address a limitation of the GMM, Bayesian Inference can be incorporated to determine the appropriate number of components in a validated manner. Bayesian Inference allows for the inclusion of prior distributions that incorporate prior beliefs or expectations about the parameters. By incorporating prior distributions, overfitting can be mitigated, leading to improved regularization [61]. Moreover, GMM enables automatic determination of the number of components, providing a more robust solution.

However, using Bayesian Inference does not fix all the limitations and even introduces another one. For instance, the constraint on the maximum number of selectable features remains. To address this subjective task of identifying the appropriate value for $K$, $K$ is determined based on the number of features that collectively account for a cumulative variance of 95%. Additionally, the incorporation of Bayesian Inference introduces further computational complexity and potential bias. To assess the impact of incorporating Bayesian Inference on the data utilized in this paper, the distribution of the data after feature selection is examined. Fig. 2C illustrates this analysis for the BRCA dataset, and it can be inferred that the distribution remains relatively unchanged compared to Fig. 2B. Both distributions still exhibit a normal distribution pattern, suggesting that these limitations do not have a significant impact. Therefore, GMM with Bayesian Inference is incorporated in the LIDAF method and the impact on the complexity of this method is shown in the fifth column of Table I.

### D. FINDING RELATIONSHIPS BETWEEN OMICS TYPES
To address another limitation of the MDICC method, linear relationships are captured between expressions of omics data pertaining to the interaction between different modules [23].

TABLE I
TCGA DATASETS SIZES DURING PRE-PROCESSING

| Dataset | Omics Type | Original | After Removal of Missing Values | After Variable Selection |
|---|---|---|---|---|
| BRCA (323) | Gene expression | 60,483 | 25,387 | 19,010 |
| | miRNA | 1,881 | 340 | 308 |
| | Methylation | 22,533 | 22,533 | 14,064 |
| KIRC (289) | Gene expression | 58,316 | 25,850 | 15,619 |
| | miRNA | 1,879 | 347 | 226 |
| | Methylation | 22,928 | 22,928 | 16,802 |
| LUAD (465) | Gene expression | 60,483 | 25,472 | 18,522 |
| | miRNA | 1,881 | 435 | 374 |
| | Methylation | 22,123 | 22,123 | 17,373 |
| LIHC (378) | Gene expression | 29,531 | 15,583 | 10,391 |
| | miRNA | 1,046 | 361 | 222 |
| | Methylation | 5,000 | 5,000 | 4,599 |
| GBM (271) | Gene expression | 12,042 | 12,042 | 10,101 |
| | miRNA | 534 | 534 | 409 |
| | Methylation | 5,000 | 5,000 | 4,095 |
| LUSC (335) | Gene expression | 20,531 | 16,858 | 11,227 |
| | miRNA | 1,046 | 396 | 253 |
| | Methylation | 5,000 | 5,000 | 4,355 |
| SARC (261) | Gene expression | 20,531 | 16,026 | 10,336 |
| | miRNA | 1,046 | 333 | 224 |
| | Methylation | 5,000 | 5,000 | 4,563 |
| OV (290) | Gene expression | 20,531 | 16,851 | 11,655 |
| | miRNA | 705 | 328 | 212 |
| | Methylation | 5,000 | 5,000 | 4,188 |
| AML (106) | Gene expression | 20,531 | 14,949 | 10,679 |
| | miRNA | 705 | 245 | 173 |
| | Methylation | 5,000 | 5,000 | 4,482 |
| SKCM (439) | Gene expression | 20,531 | 15,955 | 10,128 |
| | miRNA | 1,046 | 395 | 262 |
| | Methylation | 5,000 | 5,000 | 4,525 |

CCA is used to do this. CCA is a statistical method employed to investigate the relationship between two sets of variables. It aims to identify the linear combinations of variables from each set that exhibit the highest correlation with one another [62,63].

There are various approaches to perform CCA, with one commonly used method being Singular Value Decomposition (SVD). SVD is preferred because it can handle non-square variable sets, produces interpretable outputs, and is a widely recognized and established technique [64]. SVD is a technique that decomposes a matrix $A$ into the product of three matrices: $U\Sigma V^T$, where $U$ and $V^T$ are unitary matrices and $\Sigma$ is a diagonal matrix [65]. The diagonal elements of $\Sigma$ represent the singular values of $A$.

By considering the rank of $\Sigma$, denoted as $r$, the number of most significant canonical correlations for a given number $n_{CCA}$ of canonical variates can be determined. The value $n_{CCA}$ is determined by the sample size of $X$ and $Y$, where $X$ is the predictor variable set and $Y$ the response variable set. To obtain the canonical variates for $X$, we select the first $r$ columns from $U$. Similarly, for $Y$, the first $r$ columns from $V^T$ can be taken. These selected columns represent the canonical variates for $X$ and $Y$ based on the most significant canonical correlations [66]. Subsequently, the Euclidean distance (Equation 7) is applied to calculate the distance between each pair of canonical variates of $X$ and $Y$, which is then stored in a distance matrix. Next the distance matrix is transformed into an affinity matrix (Equation 3). This transformation is carried out in six distinct ways:

1. miRNA as the predictor dataset and gene expression as the response dataset;
2. Gene expression as the predictor dataset and miRNA as the response dataset;
3. miRNA as the predictor dataset and methylation as the response dataset;
4. Methylation as the predictor dataset and miRNA as the response dataset;
5. Gene expression as the predictor dataset and methylation as the response dataset;
6. Methylation dataset as the predictor dataset and gene expression as the response dataset;

By exploring these six variations, the linear relationships between the different omics types can be represented.

CCA is chosen because it is an accessible method that can effectively uncover interactions between omics data [64]. However, it is important to consider certain aspects when applying CCA. The assumptions underlying CCA are similar to those of other general linear model techniques, such as multiple correlation, regression, and factor analysis [67]. Two commonly assumed, but not always necessary, conditions for CCA are normality and linearity in data. It is beneficial to assess whether these assumptions hold to ensure the validity of the analysis. Another factor that should be considered is that the relationship between the predictor and response datasets are not interchangeable. CCA provides valuable insights into how the response variables relate to the predictor variables. However, it does not necessarily provide insights into reverse relationships, namely how the predictor variables relate to the response variables. By considering these factors, CCA can be used effectively while being aware of its underlying assumptions and limitations. Therefore, CCA can be used as a technique in exploring relationships between omics datasets.

### E. FUSION

As mentioned above, there are six affinity matrices that need to be used for data integration. However, directly combining these affinity matrices with the ones from the CCA cannot yield satisfactory results. This is because the CCA affinity matrices are constructed differently compared to the three affinity matrices created in [23]. Moreover, the presence of six additional matrices could potentially overshadow the importance of the original three matrices. To address this, the decision is made to initially fuse each view separately. This results in two remaining affinity matrices that capture distinct perspectives of the omics data. By subsequently fusing these two matrices, a single remaining affinity matrix is obtained. This final matrix can then be utilized for clustering and survival analysis. Nevertheless, one challenge associated with this approach arises in the first and second fusion step, primarily related to the computational inefficiency in determining the suitable value for $k_2$. $k_2$ dictates the number of neighbours used to calculate $\gamma$ (Equation 5), which in turn determines the regularization strength in the fusion pipeline. To tackle this challenge, the optimal $\gamma$ is determined by performing an exhaustive search. In the first fusion step (C.1 in Fig. 1) an exhaustive search over the range $\gamma \in [2,100]$ is performed and in the second fusion step (C.2 in Fig. 1) this is over the range $\gamma \in [2, \#samples + 2]$. This parameter, $\gamma$, will be used to maximize Equation 11:

$$k_2 = \operatorname{argmax}(\overline{rr}) \text{ where } rr = \frac{i \cdot d_{i+1} - \sum_{i=2}^{N} d_{1,\dots,i}}{2}, \quad (11)$$

and where $d$ is all the rows of $(i + 2 - 2) = i$ samples in the distance matrix. Note that Equation 11 is solely applied during the first two fusion steps due to computational costs associated with exhaustive $k_2$ search for the entire fusion pipeline. However, for the last fusion step an exhaustive search is conducted in a range of $k_2 \in [2,100]$ for the entire fusion pipeline. An alternative approach to addressing the challenge of determining the optimal $k_2$ for each fusion step is to utilized clustering metrics, such as supervised metrics like the ARI and NMI score. These metrics can be used to select the $k_2$ values in the first two fusion steps that demonstrates the highest clustering performance.

Furthermore, the selection of the initial $C$ eigenvectors, which represent the $C$ associated largest eigenvalues in the fusion process, where $C$ corresponds to the number of clusters to be identified in the analysis. However, an exception is made when identifying two clusters, which leads to $C = 3$. Two matrices will remain from the initial two fusion steps. These will be scaled based on Equation 3 and fused once again using the same method used by [23]. However, this time the optimal value for $k_2$ is determined through an exhaustive search for $k_2$ ϵ [2,100] for the entire fusion method, rather than solely for Equation 11. This culminates in the production of the final matrix, referred to $S$.

## F. CLUSTRING

Eventually the fused matrix $S$ will be clustered by K-means++ where the centroids are determined by the Euclidean distance (Equation 7) and are evaluated by the Adjusted Rand Index (ARI) and the Normalized Mutual Information (NMI). The ARI extends the Rand Index (RI) proposed by [68], to measure similarity between two clusters. The ARI adjusts the RI by considering the expected agreement under a null model of randomness, effectively correcting for chance. The ARI ranges from -1 to 1, with 1 indicating perfect agreement, 0 indicating random agreement, and negative values indicating disagreement beyond chance [69,70].

The NMI is derived from the Mutual Information (MI), a measure used to quantify shared statistical information between two distributions [71]. The NMI represents a normalized version of the MI that produces scores with boundaries between 0 and 1, where a 0 implies no mutual information, while a score of 1 represents a perfect match [44]. When measuring these metrics between two clusters, one cluster is assigned as $A$ and the other as $B$. The Adjusted Rand Index can be computed as [72]:

$$ARI = \frac{2(ad - bc)}{(a+b)(b+d) + (a+c)(c+d)}. \quad (12)$$

The variables are defined as follows:
- $a$ represent the number of objects in a pair that are placed in the same group in both $A$ and $B$.
- $b$ represents the number of objects in a pair that are in the same group in $A$ but a different group in $B$.
- $c$ represents the number of objects in a pair that are in the same group in $B$ but a different group in $A$.
- $d$ represents the number of objects in a pair that are placed in different groups in both $A$ and $B$.

The NMI can be computed as [73]:

$$MI = \frac{I(A,B)}{\max(H(A), H(B))}, \quad (13)$$

where:

$$I(A,B) = \sum_{n=1}^{N} \sum_{m=1}^{M} \frac{|A_n \cap B_m|}{K} \log \frac{K|A_n \cap B_m|}{|A_n| \cdot |B_m|},$$

and:

$$H(A) = -\sum_{n=1}^{N} \frac{|A_n|}{K} \log \frac{|A_n|}{K},$$

and:

$$H(B) = -\sum_{m=1}^{m} \frac{|B_m|}{K} \log \frac{|B_m|}{K}.$$

For these equations $N$ is the number of clusters in $A$ and $M$ is the number of clusters in $B$.

---

**Algorithm 1: LIDAF**

**Input**: Multi-omics data matrices $\{X_i\}_{i=1}^{m}$;
Parameters $k_1, k_2, k_3$ and $C$.

1. **Pre-processing:**
   1) Deleting rows with more than 20% of zeros;
   2) Imputation of values;
   3) Scaling;
   4) Centring;
   5) Standardization;
   6) Power transformation;
   7) Feature selection by Bayesian Gaussian Mixture Model of $k$ features, where k is based on the number of features needed to explain the cumulative variance of 95% of the dataset;

2. **Get relationships between expressions of omics data:**
   $R_{(m-1)\cdot m} = \{X_{ij}\}_{i=1, j=i+1,\ldots,m}^{m}, \{X_{ij}\}_{i=m, j=i-1,\ldots,1}^{1}$
   **for** relation $r \in \{1 \ldots R_{(m-1)\cdot m}\}$ **do**
      $x = r_0$;
      $y = r_1$;
      Centre $x, y$;
      $U, D, V = svd(x, y)$;
      $k = Rank(D)$;
      Select $k$ significant canonical correlations;
      $W_x$ = canonical variates for $r_0$;
      $W_y$ = canonical variates for $r_1$;
      $n = length(W_x)$
      Create a distance matrix based on the Euclidean distance:
      $d_{ij} = \sqrt{(W_{xi} - W_{xj})^2 + (W_{yi} - W_{yj})^2}$ where $i = 1,\ldots,n$, $j = i+1,\ldots,n$;
      $d_{ij} = \sqrt{(W_{xi} - W_{xj})^2 + (W_{yi} - W_{yj})^2}$ where $i = 1,\ldots,n$, $j = i+1,\ldots,n$;
   **end**

3. **Fusion:**
   Iterating over $i \in [2,100]$ as parameter in the fusion function for the affinity matrices gathered from inter dataset relationships;
      Use fused matrix that maximizes rr;
   Iterating over $j \in [2, \#\text{number of samples} - 2]$ as parameter in the fusion function for the affinity matrices gathered from CCA;
      Use fused matrix that maximizes rr;
   Save matrices with the optimal $W$ in list $M$ and $j$ in list $Z$
   Save combination in $W$ and $Z$ in a list $Q$;
   Iterating over $n \in [2,100]$ as parameter for $k_2$ in the fusion function for the affinity matrices gathered from $Q$;

**Output**: 1 fused matrix $S$ ready for K-means++ and survival analysis

## IV. RESULTS

In this section, a comparative analysis will be conducted between the proposed LIDAF method and MDICC and three other state-of-the-art clustering methods used for identifying cancer subtypes which were able to outperform MDICC in some survival analysis instances: MSCA [72], SNF [28], and ANF [74]. The aim is to assess their respective performances and evaluate their effectiveness. This will be evaluated based on the following classification of thresholds [72, 75]:

- Excellent performance: ARI and NMI ≥ 0.90;
- Good performance: 0.80 ≤ ARI and NMI < 0.90;
- Moderate performance: 0.60 ≤ ARI and NMI < 0.80;
- Poor performance: ARI and NMI < 0.60.

### A. CLUSTERING EVALUATION

In the clustering evaluation of the four cancer types BRCA, LIHC, LUAD and KIRC, a comparison was made between the MDICC method, and the proposed LIDAF presented in this paper. The findings will be explained in this section.

For the BRCA cancer type, the MDICC method demonstrated ARI results over 0.90 in 83 out of 99 cases (Fig. 3), and its NMI values were higher than 0.80 in 83 out of 99 cases (Fig. 4). On the other hand, the proposed LIDAF showed ARI values over 0.90 in 86 cases and NMI values over 0.80 in 86 cases.

In terms of the KIRC cancer type, MDICC performed ARI values over 0.90 for 77 $k_2$ values and NMI values over 0.80 in the same number of cases. The proposed LIDAF method achieved for ARI values over 0.90 and NMI values over 0.80 for each $k_2$ value. However, for LIHC and LUAD, both MDICC and the proposed LIDAF fell below the threshold of excellence for both the ARI and NMI values. Nonetheless, the proposed LIDAF displayed a higher number of ARI and NMI values above the moderate threshold for these cancer types compared to the MDICC method. Specifically, when examining LUAD, the proposed LIDAF outperformed MDICC with 48 higher ARI values, while MDICC only had 10 higher ARI values. Conversely, for LIHC, the proposed LIDAF had 27 higher ARI values compared to MDICC's 48 higher ARI values.

An important observation is that even when the ARI values of the proposed LIDAF are lower than those of the MDICC method, they still fall within the moderate range (and even close to good). On the contrary, the ARI values of the MDICC method are consistently very poor when compared to the proposed LIDAF. This pattern can also be recognized for the NMI values of LIHC.

Furthermore, there is an unresolved issue with an awkward minimum ARI, resulting in a negative ARI, for the proposed LIDAF at $k_2 \in [59,60]$, and $k_2 = 65$ for the MDICC method. This could possibly be attributed to the 59$^{th}$ and 60$^{th}$ closest neighbours being ambiguous and significantly influencing the lack of similarity. Furthermore, on BRCA, the proposed LIDAF has 16 ARI values higher than MDICC, while MDICC has 31 higher ARI values. However, the difference in ARI values is lower for the proposed LIDAF compared to MDICC. In fact, the proposed LIDAF method has three more ARI values above 0.90 than MDICC. The results for KIRC with the proposed LIDAF are nearly perfect, with all ARI values surpassing the threshold of 0.90.

Additionally, the proposed LIDAF has higher ARI values for 33 $k_2$ values compared to 20 $k_2$ values for MDICC. In general, the proposed LIDAF demonstrates superior performance compared to MDICC method for each four cancer types when considering the number of ARI values that can be categorized as moderate to excellent. A recurring pattern observed is that lower values of $k_2$ tend to correspond to higher ARI values for the proposed LIDAF compared to the MDICC method. According to these findings, LIDAF exhibits more robustness across various $k_2$ values when compared to MDICC.

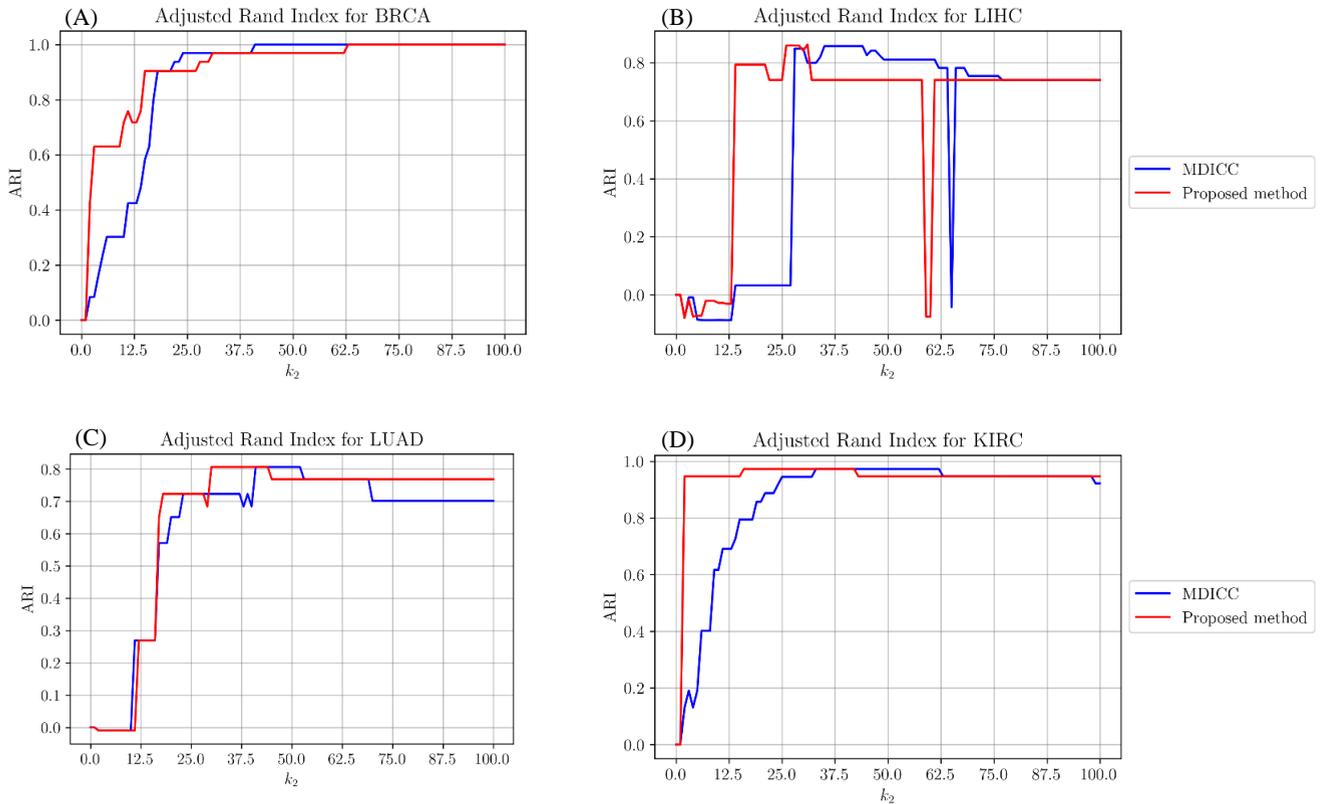

**FIGURE 3.** Each graph is based on analysis with a clustering which uses K-means++ with the Euclidean distance for 2 clusters. (A) ARI distribution for different values of $k_2$ for the proposed LIDAF method and the MDICC method for BRCA; (B) ARI distribution for different values of $k_2$ for the proposed LIDAF method and the MDICC method for LIHC; (C) ARI distribution for different values of $k_2$ for the proposed LIDAF method and the MDICC method for LUAD; (D) ARI distribution for different values of $k_2$ for the proposed LIDAF method and the MDICC method for KIRC.

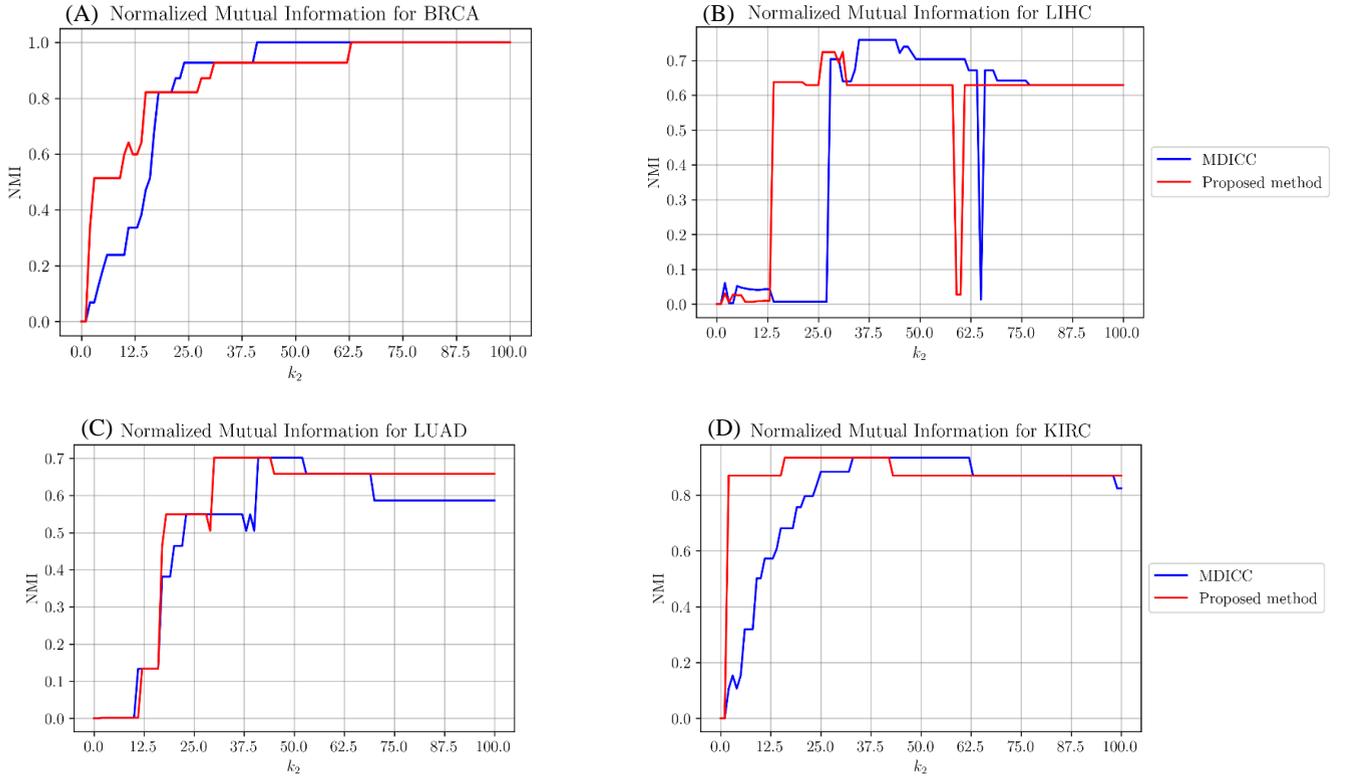

**FIGURE 4.** Each graph is based on analysis with a clustering which uses K-means++ with the Euclidean distance for 2 clusters. (A) NMI distribution for different values of $k_2$ for the proposed LIDAF and the MDICC method for BRCA; (B) NMI distribution for different values of $k_2$ for the proposed LIDAF and the MDICC method for LIHC; (C) NMI distribution for different values of $k_2$ for the proposed LIDAF and the MDICC method for LUAD; (D) NMI distribution for different values of $k_2$ for the proposed LIDAF and the MDICC method for KIRC.

Fig. 4 provides an overview of the NMI distribution for both the proposed LIDAF and the MDICC method. When considering BRCA, the MDICC method outperforms the proposed LIDAF in terms of NMI values, with 31 $k_2$ values higher compared to 16 for the proposed LIDAF. Additionally, for a NMI score above 0.90, the MDICC method slightly outperforms the proposed LIDAF, with 77 values above 0.90 compared to 70. However, when using a threshold of 0.80, which represents a good NMI score, the proposed LIDAF has 86 values above this threshold, while the MDICC method has 83 values. A comparable trend is noticeable in the case of LIHC, wherein the MDICC method exhibits more numbers of favourable NMI values when the threshold is set to 0.70 (since there are no NMI values exceeding 0.80). On the other hand, when the threshold is set to 0.60, the proposed LIDAF shows a higher count of moderate NMI values, with 85 compared to 72. For the KIRC and LUAD cancer types, similar patterns can be observed. The proposed LIDAF consistently achieves NMI values higher than 0.80 for all $k_2$ values, indicating an almost excellent score. On the other hand, the MDICC method has 77 values for $k_2$ that are above 0.80.

Furthermore, for KIRC, the proposed LIDAF has 33 values higher than MDICC, while MDICC has 20 values higher than the proposed LIDAF. In the case of LUAD, the proposed LIDAF has 48 higher values compared to the MDICC method, while MDICC has 10 higher values than the proposed LIDAF. Moreover, the proposed LIDAF obtains 71 moderate NMI values, whereas the MDICC method has 29 values for $k_2$ that can be considered moderate. These results once again demonstrate the enhanced robustness of the proposed LIDAF when evaluated across a range of $k_2$ values, in comparison to MDICC.

### B. SURVIVAL ANALYSIS EVALUATION

In addition to assessing clustering performance, this paper also incorporates survival analysis to enhance the understanding of how well LIDAF identifies cancer subtypes. This analysis encompasses all ten cancer types detailed in the "Methods" section. For each cancer type, the event of interest is defined as a Boolean variable, "Death", and the time until this event occurs is represented by an integer value, "Survival Time".

For each cancer type, LIDAF partitions the fused matrix $S$ from Fig. 1C into 3,4 and 5 clusters. From the survival data, we compute the $-\log_{10}$ rank p-values using Cox proportional hazard regression. A significant $-\log_{10}$ rank p-value suggest that the clustering approach not only distinguishes clusters but also captures variations in survival outcomes among the identified cancer subtypes.

To assess the performance of LIDAF, we compare our proposed method to four state-of-the-art clustering methods: MDICC, MSCA, SNF, and ANF. Notably, MSCA, SNF, and ANF exhibit superior performance compared to MDICC in specific scenarios. MSCA, SNF, and ANF were able to outperform MDICC in certain scenarios, such as KIRC with 3 and 4 clusters, LUAD with 3 clusters, LIHC with 4 and 5 clusters, GBM with 3 clusters, LUSC with 4 and 5 clusters,

GBM with 3 clusters, LUSC with 4 and 5 clusters, and SKCM with 3, 4, and 5 clusters. In other words, 19 out of 30 p-values from the MDICC method were more significant compared to the other three state-of-the-art clustering methods.

The p-values obtained from our proposed method LIDAF, as shown in Table II, has instances where the proposed LIDAF achieved more significant p-values compared to MDICC. 50% of the p-values for the proposed LIDAF are more significant compared to the p-values from the MDICC method.

Moreover, when compared to four state-of-the-art clustering methods, LIDAF exhibits the highest number of most significant p-values. Table II highlights these findings with ANF displaying the highest p-value among the other methods just once, followed by MSCA with four instances, SNF with six, MDICC with nine, and our proposed method, LIDAF, leading with ten. Notably, LIDAF demonstrated improvement in over fifteen p-values compared to MDICC. This improvement is evident in Table II, where LIDAF outperformed MDICC in identifying cancer subtypes based on survival analysis, particularly for BRCA, GBM, and AML when clustering with 4 clusters, KIRC with 5 clusters, LUAD and LIHC with 3, 4, and 5 clusters, SARC and SKCM with 4 and 5 clusters, and OV with 3 clusters.

Nevertheless, it should be noted that the p-values for LUSC with 3, 4, and 5 clusters, and AML with 3 clusters were not found to be significant. This discrepancy could be attributed to the fact that determining the optimal parameter $k_2$ is a challenging task.

In LIDAF, the $k_2$ value was chosen based on maximizing Equation 11. However, the discovery of this $k_2$ value does not necessarily assure its optimality. To determine the optimal $k_2$ for our method, the exhaustive search used in the third fusion step should be applied. However, this is considered too computationally expensive and is therefore not implemented. A less computational expensive way to find the optimal $k_2$, which maybe could improve LIDAF, is setting the first fusion step to $k_2 = 42$, based on the results of [23], and determining the second fusion step by selecting the highest average between ARI and NMI, which could result in improved p-values on LIHC, for instance, with values of 3.80 for 3 clusters, 3.30 for 4 clusters, and 3.45 for 4 clusters, offering the best outcome among the four methods. However, basing the parameter selection solely on ARI and NMI would lead to a more supervised method, which differs from the nature of MDICC, and other p-values were less or not significant compared to the method proposed in this paper. Consequently, this alternative method is not further considered. Furthermore, optimizing the tuning parameter $C$, which signifies the number of largest eigenvalues to be selected in the fusion process, has the potential to enhance performance. For example, setting $C = 4$ for AML yields a significant p-value for AML with 3 clusters.

## V. CONCLUSION

This paper introduces a novel approach called LIDAF, designed for the integration of multi-omics data in cancer analysis. The method entails constructing affinity matrices that capture linear relationships between distinct types of omics data using CCA. Also, affinity matrices are constructed that capture relationships within each omics type and between individual samples. The pipeline involves a series of steps: addressing missing values and zeros, imputing data, applying Z-score standardization, performing the Yeo Johnson transformation, selecting variables through Bayesian Gaussian Mixture model, and employing CCA.

LIDAF enhances its capabilities by incorporating these affinity matrices and fusing them in three sequential steps. In the course of this modification, the paper introduces a hypothesis in the introduction section. Based on the evaluation of clustering performance, it becomes evident that the null hypothesis can be rejected. This claim is supported by the observation that LIDAF exhibit more robustness in terms of clustering performance when compared to MDICC. This conclusion is drawn from its good performance at lower $k_2$ values and its ability to achieve comparable performance at similar $k_2$ values.

TABLE II
SURVIVAL ANALYSIS ON THE TCGA DATASETS USING THE PROPOSED LIDAF AND MSCA, SNF, ANF, AND MDICC METHODS. $k_3$ IS THE NUMBER OF CLUSTERS USED AND ALL THE OTHER VALUES ARE DETERMINED BY USING -$\log_{10}$ P-VALUES FROM THE LOG-RANK TEST, WHERE THE BEST RESULT IS INDICATED IN BOLD.

| Cancer type | $k_3$ | MSCA | SNF | ANF | MDICC | LIDAF (proposed) |
|---|---|---|---|---|---|---|
| BRCA | 3 | 2.31 | 2.40 | 1.87 | **9.87** | 4.13 |
|  | 4 | 2.80 | 2.00 | 1.48 | 5.30 | **5.76** |
|  | 5 | 3.01 | 1.70 | 1.52 | **13.19** | 6.31 |
| KIRC | 3 | 3.55 | **8.30** | 3.12 | 3.56 | 2.10 |
|  | 4 | 3.42 | **7.63** | 3.22 | 3.09 | 2.86 |
|  | 5 | 2.16 | 2.86 | 2.47 | 4.48 | **4.65** |
| LUAD | 3 | **2.55** | 0.35 | 0.64 | 1.48 | 2.19 |
|  | 4 | 1.62 | 2.12 | 0.66 | 2.53 | **3.14** |
|  | 5 | 2.72 | 2.57 | 1.08 | 3.56 | **6.97** |
| LIHC | 3 | 0.94 | 0.68 | 0.99 | 1.87 | **1.95** |
|  | 4 | 0.58 | **2.05** | 0.72 | 1.91 | 1.93 |
|  | 5 | 0.93 | **2.47** | 0.67 | 1.57 | 1.94 |
| GBM | 3 | 2.27 | 0.47 | **4.69** | 4.13 | 3.32 |
|  | 4 | 1.62 | 2.74 | 3.23 | 4.07 | **6.63** |
|  | 5 | 3.60 | 3.37 | 3.69 | **4.16** | 2.88 |
| LUSC | 3 | 0.74 | 0.03 | 0.56 | **1.58** | 0.86 |
|  | 4 | 1.38 | **2.11** | 0.35 | 1.65 | 1.15 |
|  | 5 | 2.14 | **2.15** | 0.78 | 1.45 | 0.84 |
| SARC | 3 | 1.22 | 2.17 | 1.16 | **4.60** | 4.56 |
|  | 4 | 0.89 | 1.80 | 1.00 | 4.19 | **5.65** |
|  | 5 | 1.03 | 3.26 | 0.76 | 5.00 | **21.98** |
| OV | 3 | 0.67 | 0.27 | 1.70 | 4.77 | **5.41** |
|  | 4 | 0.37 | 0.48 | 0.82 | **4.92** | 4.82 |
|  | 5 | 0.66 | 2.34 | 0.40 | **5.89** | 4.62 |
| AML | 3 | 1.36 | 1.00 | 0.79 | **2.24** | 0.38 |
|  | 4 | 1.33 | 2.26 | 1.48 | 2.61 | **3.01** |
|  | 5 | 1.49 | 2.66 | 1.41 | **3.50** | 3.21 |
| SKCM | 3 | **5.34** | 0.47 | 3.54 | 1.61 | 1.53 |
|  | 4 | **3.38** | 0.18 | 2.60 | 1.57 | 2.15 |
|  | 5 | **6.15** | 0.10 | 2.03 | 1.67 | 1.71 |

Additionally, the null hypothesis is rejected for 50% of the -$\log_{10}$ rank p-values obtained from survival analysis compared to [23]. Furthermore, LIDAF possesses the highest number of p-values that exhibit greater significance compared to the others. However, it is essential to acknowledge that the rejection of $H_0$ may vary depending specific cancer types and cluster investigated in this paper.

In summary, LIDAF can yield desired results in identifying linear relationships among different types of omics data while simultaneously mitigating computational complexity and addressing the heterogeneity often present in omics data through its pre-processing pipeline. Furthermore, LIDAF serves as a method capable of harmonizing diverse views on multi-omics data through a three-step fusion pipeline, where relationships between samples and among different omics types are given equal importance. These findings suggest that considering both the affinity between various omics types and the relationship among patients within the same omics types, which addresses the limitations of MDICC, can lead to maintenance or even improved performance when working with multi-omics data.

## VI. DISCUSSION

While the proposed LIDAF in this paper shows improvement in clustering, in half of the -$\log_{10}$ rank p-value and addresses the limitations of the MDICC method, it also has some suggestions for future research. Firstly, the subjective nature of setting parameters for handling missing values, imputing values, and feature selection poses a challenge. Further research could explore more robust and objective approaches to overcome this limitation. Additionally, the use of SVD for finding linear combinations, despite its interpretability and ability to handle non-square sets of variables, is computationally expensive. To mitigate this, the optimal $k_2$ is determined through an exhaustive search, tuning parameters based on maximizing Equation 11 instead of the whole fusion pipeline. However, this approach considers only one value for $k_2$, potentially overlooking other values that could yield comparable results. Future work could incorporate additional metrics, such as determining $k_2$ on clustering metrics, such as the ARI or NMI to determine the optimal $k_2$ at each fusion step or another way to find the optimal $k_2$.

Also, it should be acknowledged that the MDICC method and the proposed LIDAF in this paper only utilize three types of omics data. However, it is plausible that other types of omics data may exhibit more favourable linear relationships, resulting in improved affinity matrices based on CCA. However, the proposed LIDAF bases it relationships on CCA which only considers linear relationships between omics datasets. It is worth noting that the interaction between omics data can be complex and non-linear, and thus, exploring alternative methods for identifying relationships between omics datasets while using the pipeline considered in this paper could still potentially lead to even better performance. Therefore, future research could consider other techniques that capture non-linear relationships.

Moreover, by utilizing affinity matrices, our proposed method exhibits adaptability beyond the domain of cancer identification. To illustrate this versatility, consider a study from [76], which delves into interval uncertainty in game theory. This study defines a cooperative interval game as an ordered pair $< N, w >$, where $N$ represents the set of players and $w$ their characters. Each player, denoted as $n$, forms coalitions, denoted as $S$, based on specific criteria. In this context, LIDAF can be utilized by uncovering similar strategies based on the characteristics of $w$. This is achieved by identifying similarities among these characteristics, ideally expressed through a distance metric, both among players and within the intervals of $w(S)$ for each player. Another application for LIDAF is to enhance the comprehension of the relationship between economics and health activities, which can have profound implications for policymaking. For instance, in [77], a hierarchical clustering method, factoring in elements such as Gross Domestic Product per capita, industrial production, COVID-19 cases per population, patient recovery rates, COVID-19 death cases, and more. This approach aids policymakers in gauging economic indices in relation to these considered factors, thereby identifying specific influences on economic developments. By generating distance matrices that capture similarities among these factors across countries and distances between factors within each country, LIDAF can be employed to cluster countries into distinct groups.

In summary, while the proposed LIDAF method had demonstrated almost equal or even better improvements in clustering and survival analysis while addressing the limitations of the MDICC framework, it also points the way forward for future research. Further exploration can be done towards challenges posed by subjective parameter settings, optimizing computational efficiency in linear combinations, exploring non-linear relationships among omics datasets, and expanding its application into diverse domains.

## VII. CODE AVAILABILITY

The code utilized in this paper can be accessed at the following GitHub repository:
https://github.com/Peelen-Mark/identifying-cancer-subtypes-code

**Mark Peelen**, did his Bachelor thesis for his B.Sc. degree with honor in Artificial Intelligence from Radboud University Nijmegen in 2023. Where he will continue his interest by starting his M.Sc in Artificial Intelligence: Intelligent Technology in 2024. His research interests in the area of AI in healthcare include data fusion of multi-omics data for cancer subtype diagnosis.

**Leila Bagheriye** (M20) received the B.Sc. degree in electrical engineering from the University of Tabriz, Tabriz, Iran, in 2010 and the M.A.Sc. degree in electronics engineering with honor from the University of Zanjan, Zanjan, Iran, in 2012, She received her Ph.D. degree in electronics engineering in 2018 and she was a visiting Ph.D. student at the ICE-Laboratory, Aarhus University, Aarhus, Denmark. From 2019-2021 she was with the CAES group at the University of Twente and focused on machine learning based reliability enhancement of multi-sensor platforms. Since 2021 she is with the Donders Institute at Radboud University where she focuses on implementation of neuromorphic algorithms emphasizing on Bayesian networks and spiking neural networks for cancer subtype diagnosis.

**Johan Kwisthout** is associate professor in Artificial Intelligence and principal investigator at the Donders Institute at Radboud University, where he leads the Foundations of natural and stochastic computing group. He holds M.Sc. degrees in computer science and artificial intelligence and received his Ph.D. degree in computer science in 2009 from Utrecht University. He is interested in the foundations and applications of probabilistic graphical models, particularly computations on Bayesian networks, as well as their realization in neuromorphic architecture.